# Big Data Analytics Applying the Fusion Approach of Multicriteria Decision Making with Deep Learning Algorithms

Swarajya Lakshmi V Papineni[1], Snigdha Yarlagadda[2], Harita Akkineni[3], A. Mallikarjuna Reddy[4]

[1]Professor, Department of Information Technology, Prasad V Potluri Siddhartha Institute of Technology, Vijayawada, AP, India

[2] Technology Analyst, Infosys Limited, Bangalore, Karnataka, India

[3]Associate Professor, Department of Information Technology, Prasad V Potluri Siddhartha Institute of Technology, Vijayawada, AP, India

[4]Assisstant Professor, Department of CSE, Anurag University, Hyderabad, AP, India

[1]papinenivsl@gmail.com, [2] snigdha.y928@gmail.com, [3]akkinenih@gmail.com, [4]mallikarjunreddycse@cvsr.ac.in

***Abstract:*** *Data is evolving with the rapid progress of population and communication for various types of devices such as networks, cloud computing, Internet of Things (IoT), actuators, and sensors. The increment of data and communication content goes with the equivalence of velocity, speed, size, and value to provide the useful and meaningful knowledge that helps to solve the future challenging tasks and latest issues. Besides, multicriteria based decision making is one of the key issues to solve for various issues related to the alternative effects in big data analysis. It tends to find a solution based on the latest machine learning techniques that include algorithms like decision making and deep learning mechanism based on multicriteria in providing insights to big data. On the other hand, the derivations are made for it to go with the approximations to increase the duality of runtime and improve the entire system's potentiality and efficacy. In essence, several fields, including business, agriculture, information technology, and computer science, use deep learning and multicriteria-based decision-making problems. This paper aims to provide various applications that involve the concepts of deep learning techniques and exploiting the multicriteria approaches for issues that are facing in big data analytics by proposing new studies with the fusion approaches of data-driven techniques.*

## I. INTRODUCTION

From the analysis of big data, the major challenge is to manage the unrelated set of data to be converted on a single basis [1]. The conventional methods were failed to do so the process for the approach to provide the correct insight for the data ingestion that includes the data division based on the velocity, size, and in different shapes or formats which are to be further intended to be responsive enough for it to be applicable for the number of optional applications, with this progress of population and the communication field of different kinds of things like networking, soft computing, IoT, and variety of knowledge shared for a meaningful understanding of data contents on the related fields in computer science, information technology, and agriculture. According to the studies, almost locally, the 1.5 ZB of data [2] has been created that made the challenges in case of information for insight knowledge on the researchers' issues. In the global view, it has been decremented to 1.3 ZB for it to prepare easy as on the data created and for the utilization of applications beyond the real-time environment. There are various tools and algorithms to solve and handle the large set of data called to be the big data [3] in an effective manner for the electronic data available from various sources. By the end of 2020, the data that can be created can be approximating at the rate of 50 ZB. These developed tools are capable of achieving the process of the acquired data, later, applying various [4] filtering technique for the process to be done for the data to convert in an understandable meaning and to find proper insights to communicate the output with external tools under the ongoing process for data to be exchangeable.

For the process of making the data to be understandable, various machine learning (ML) algorithms were applicable, such as neural networks, gravitational search, support vector machine, tabu search, particle swarm optimization (PSO), long short term memory network (LSTM), genetic algorithm, fuzzy logic, and deep learning techniques. It is all related to the data-driven approach for it to be provided with a large sequence of data to understand the process of data for training the network. Combining the decision-making analysis makes the process entirely multiple criteria under technical issues for providing the researchers with a concrete idea for future development in the analysis of the big data sets. Besides, the processing involves the advantageous out of making it possible for application in different fields for better choice-making, effective and efficiency overall in enhancing the potentiality for the practitioner of the support [5],[6] making system in finding the novelty in the application of various domains involved.

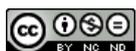

Various decision-making support systems exist in the real-world environment that improve potentiality in achieving this concerning issue [7]. Although, decision seeking issue plays a crucial step in the successful outcome of an institution or organization. Depending on the multiple criteria, the provided difficulties are further available to handle the proper data acquisition aspect. In recent times, the problem solving on big data is up to the benchmark [8] by processing the output. Still, they did not facilitate the deep learning technique to attain more intellectual decision support on the entire system based integration. This paper studies the combined analysis of data collected [9], and processing is based on the proposed novel fusion approach by integrating deep learning with the multi-criterion decision-making problem.

## II. Existing Methods

In the existing methods, it is found that the recent findings on the approaches to solve the issues and challenges behind the big data analysis to achieve precision on the processing of data and acquiring it to further applications for the other algorithms to the exploitation of support-oriented systems. A single based deep learning (DL) model with the integration of shallow kind model has been established to use t in hierarchical structural tree format. Though it is developed for the sample discretization into various levels to form the clusters, every deep learning model is trained by the related discrete model [10]. Five samples have been developed. Thus, proposed a big data hierarchical deep learning, known to be the BDHDLS, to organize the multiple criterion models. This unique study makes the distribution of cluster adapted [11], [12] for the model that has been trained. There exists five different phased. In the foremost phase, the features of behavior and contents are chosen using the tools and techniques that involve big data's objective.

Later, the second phase compares the overall data sequence on a single cluster format by doing it in a parallel improvement by the K-means approach of the cluster [13]. The hierarchical based on the clustering procedure will be added for every concentration of single cluster included with minimum excellence simultaneously in analogous with the produces a cluster of the obtained subtree. This will be fed for the final generalized approximation to get the information of achieving the greater data pattern from the cluster analysis. The estimation error can be reduced as the computing rate occurs by organizing the multiple criterion models. This unique study makes the distribution [14] of cluster adapted for deep learning schematics. Combining the decision-making analysis makes the process entirely multiple criteria under technical issues for providing the researchers with a concrete idea for future development in the analysis of the big data sets.

In Multivariate regression, we can easily find the output from the optimized theta coefficients. It is fast compared to the neural network approach since it requires less training data to model the regression analysis. K-Nearest Neighbor (KNN) is one of the supervised learning techniques in machine learning, which analyzed that for both classification basis and regression problems, we use KNN for the regression case to predict the continuous values. Our work suggests many new techniques in solving the machine algorithms oriented that introduce the deep insights of implementing the collective influences that involve the multiple existences of variables in finding a solution for the processing data found at the data collected in the system's electronic fields.

## III. Proposed Method

Large sequence data employs the deep learning-based with a fusion of multiple criteria-based decision making the hierarchical clustering algorithm for it to discover various features of which increased products hard obtaining precise characteristics for the remoteness ascertaining likeness among tests, uproarious for inclusive of unessential data at present joined group staggered progressive based on decision tree named structure. On the other hand, the unessential data significantly diminish the viability interruption discovery framework.

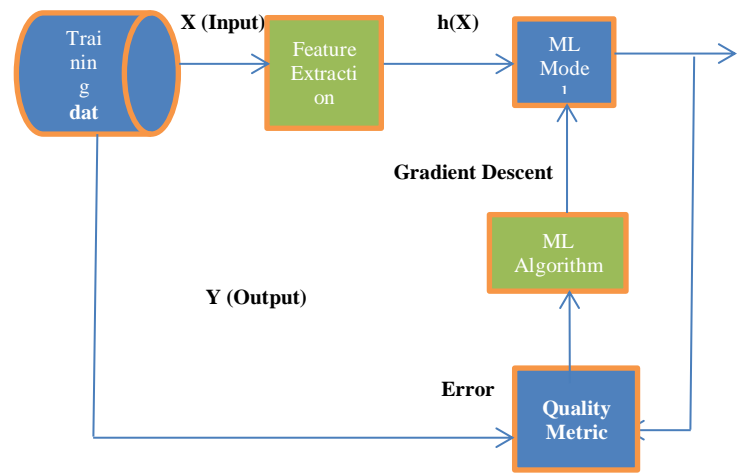

**Fig 1: Block diagram of Multiple Variate Linear Regression based deep learning algorithm**

To improve the interruption discovery execution of the progressive effects for the profoundly prepared bunch staggered group dependent mixed on conducting highlights for the substance highlights [13-14]. Various decision-making support systems exist in the real-world environment that improve potentiality in achieving this concerning issue. Since design doesn't take the structure of information into thought, it might confront trouble to deal with information with uncommon qualities. Figure 1 represents the square outline of Multiple Variate Linear Regression based ML-DL calculation. Compared to the normal, profoundly associated network, it has two significant qualities: (1) sparseness of association and (2) sharing of weight. The algorithm steps discussed below give how the end to end process is implemented.

**Algorithm steps to implement MC-DL for being used in the analytics of big data:**

1. The input of the model considered: a, b, c d model parameters defined from the datasets collected from the running records.

2. Output: Is to be observed based on the ordered list of the problem solved

3. Initializations with a batch of the data stored

4. For each dataset

    B($a_i$ k)=b($a_i$)+b($N_i$) for the entire dataset followed

    The overall effectiveness of ai's coefficient is calculated at the influencing relation between the agents of one and other initial values if they exist at the global set of problems.

    b($N_i$)= Neighboring set of values

5. The initial value that was fed to the system is expressed as,

    $\mu_{ij}$=b($a_i$)+b($a_j$)/K

    The constant normalization parameters are the beneficial agent factors, b(N)=$\sum_{j \in N} \mu_{ij} b(a_j)$

6. The mapping function defined for the entire hidden layers for the DL is

    $a_{i+1} = f_k(W_i * a_i + b_i)$ with its activation, i.e., the sigmoid function may be cumulatively added with the weights and hidden layers connecting the networking of DL.

    The decision value that is normalized for the variable of x is

    Decision value(x)=$\frac{f_k(x) - mean\ value}{\delta_k}$

7. End for
8. Return

Computing benefit factors of the neighboring agents, Mutual coefficients under the common agents can be calculated based on the weighting index.

$$\mu_{ij} = \frac{b(a_i) + b(a_j)}{k} * (C_{a_j} + 1)$$

Therefore, the overall assessment of the benefit plan will be investigated.

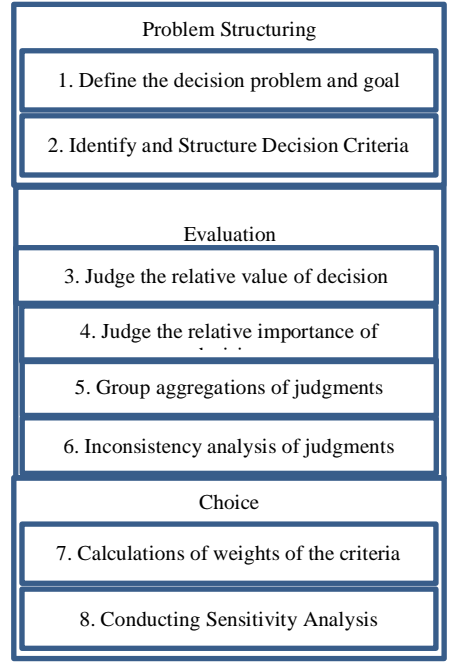

**Fig 2: Entries based on ranking for multiple criterion decision making**

Entries based on ranking for multiple criterion decision making are depicted in figure 2. Though it is developed for the sample discretization into various levels from the clusters, every model of deep learning is trained by the model on related discrete up to which five samples have been developed. Thus, the proposed big data exploits the key feature of the hierarchical oriented [15, 21-22] deep learning technique, known to be the BDHDLS, to organize the multiple criterion models [16-20]. This unique study makes the distribution of cluster adapted for the model that has been trained.

**IV. Results and Discussion**

Every model of deep learning is trained by the model on related discrete up to which five samples. The real time-based end to end process of the data that has been processed is depicted in figure basis. It attains for the start process of which it has the historical relationship for long term. Also, for real-time datasets, it will be developed on a streaming basis of a memory unit that can store the large set and finally attain into the storage part of the processing unit. For each data point p, the data points whose input variables (X) are most near to point (p) are selected.

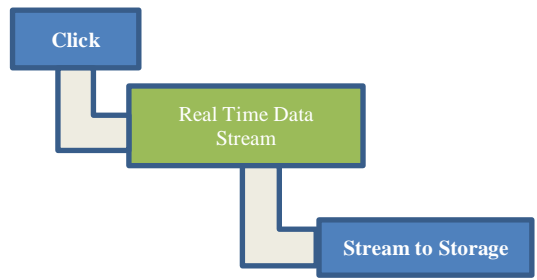

**Figure 3: End to end data processing**

Besides, it is a nonparametric supervised learning technique in machine learning applicable to regressive analysis and classification-based problems. Here we use the DL for the regression case to predict the continuous values. Here suppose we select many neighbors. For each data point p, the data points whose input variables (X) are most near to point (p) are selected. Then, distances are calculated using Euclidean distance, Manhattan, or Minkowski distances method. Then the mean of these 10 points is used as the predicted output value for p.

**Table 1: Numerical Prediction accuracy for the fields estimated**

| Techniques applied | Statistical analysis | Estimation error |
|---|---|---|
| Decision tree | -0.9560 | 1015.56 |
| KNN | -5.432 | 1867.55 |
| Ridge | -11.2645 | 986.324 |
| Linear Regression | -9.5399 | 3512.369 |
| Proposed MLDM | -0.0123 | 586.369 |

**Table 2: Categorical Prediction accuracy for the fields estimated**

| Techniques applied | Precision value | Accuracy |
|---|---|---|
| Gaussiandistribution | 0.21 | 0.13 |
| Bernoulis approximation | 0.23 | 0.49 |
| Decision tree | 0.43 | 0.98 |
| Support vector machine (SVM) | 0.68 | 0.59 |

The results obtained for the accuracy values and prediction of different fields, such as numerical and categorical, are done as per the utilization of data processed. This will solve the challenges that are faced by the system on which the data is imposed.

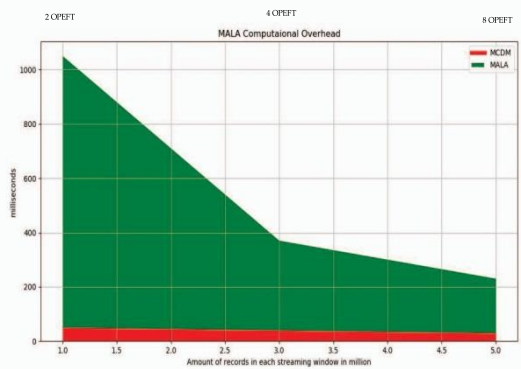

**Fig: 4 Multiagent Lamba architecture for computational overhead**

In online detection, the structural data provided will be in attaining the data acquisition termed in the interlink of property and runs through the query kind of language for better result analysis. In figure 4, the green area indicates the time of processing by the MALA. The red area indicates the latency obtained because of multicriteria decision-making. In the response, the time obtained is at the rate of average. The latency of MCDM reduced linearly with the node number of cluster increments.

Further, obtaining the window for the length of data supposed to be in the selection aspect. However, this will not enhance the faster estimation time because of the metadata availability and an extensive sequence of data with the processing data acquired in the overall factor's minimal impishness. The latency test occurred overhead to process the delay time, 4.9% rate in the average responsiveness.

## V. CONCLUSION

In this study, the present scenario of machine learning with multiple criteria decision-based problem solving has played a challenging role in increasing the population to deal with big data problems. The challenges facing the growth of volume in the data applications in various domains have been highlighted. With the framework of networking that is oriented with the machine learning algorithms, the online mode system with the equivalence rate has been attained worthy for the implementation of analytical solving. There is a trade-off between the data velocity and the processing time. Thus, attaining the cost-effective based standalone algorithms for existing benchmark results with the proposed fusion approach. The proposed study provides improved knowledge for the research on this particular stream. It supports achieving better insights in attaining the effectiveness, robustness, and accuracy in predicting the values for multiple criteria based supportive decision modeled system and their framework. Thus, providing significant improvement in the performance of advanced techniques compared to existing techniques for big data analysis through mixed data processing resources.